\documentclass[letterpaper, 10 pt, conference]{ieeeconf/ieeeconf}

\IEEEoverridecommandlockouts

\usepackage{soul}

\usepackage[per-mode=fraction]{siunitx}
\usepackage[normalem]{ulem}
\usepackage{xspace}
\usepackage{xcolor}

\DeclareSIUnit\px{px}
\DeclareSIUnit\fps{fps}

\definecolor{OliveGreen}{RGB}{0,200,25}

\newcommand{\ie}{i.\,e.\xspace}
\newcommand{\eg}{e.\,g.\xspace}
\newcommand{\etal}{et\,al.\xspace}

\newcommand{\ackHariaJubot}{The research leading to these results has received funding from the European Union’s Horizon 2021 Research and Innovation programme under grant agreement No 101070292 (HARIA) and from the Carl Zeiss Foundation through the JuBot project. }

\usepackage{amssymb}
\usepackage{amsmath}
\usepackage[USenglish]{babel}
\usepackage{bm}
\usepackage{multirow}
\usepackage{booktabs}
\usepackage[hidelinks]{hyperref}
\usepackage{cleveref}
\usepackage{enumerate}
\usepackage{graphicx}
\usepackage{pgf}
\usepackage{tikz}
\usetikzlibrary{calc}
\usepackage{ragged2e}
\usepackage{cuted}
\usepackage{subcaption}
\usepackage[font=small,labelfont=small]{caption}

\newcommand{\Exp}[2]{\text{Exp}_{#1}\left(#2\right)}
\newcommand{\Log}[2]{\text{Log}_{#1}\left(#2\right)}
\newcommand{\realR}{\mathbb{R}}
\newcommand{\expectation}[1]{\mathbb{E}\left(#1\right)}
\newcommand{\manifoldM}{\mathcal{M}}
\newcommand{\manifoldS}{\mathcal{S}}
\newcommand{\tangent}[2]{\mathcal{T}_{#2}{#1}}
\newcommand{\parallelTransport}[3]{\Gamma_{#1\rightarrow#2}\left(#3\right)}

\newcommand{\trsp}{\mathsf{T}}

\definecolor{matplotlib_red}{HTML}{D62728}
\definecolor{matplotlib_green}{HTML}{2CA02C}
\definecolor{matplotlib_blue}{HTML}{1F77B4}
\definecolor{matplotlib_black}{HTML}{000000}
\definecolor{matplotlib_orange}{HTML}{FF7E0E}

\DeclareMathOperator*{\argmax}{argmax}
\DeclareMathOperator*{\argmin}{argmin}

\title{\LARGE \bf
Incremental Learning of Full-Pose Via-Point Movement Primitives \\ on Riemannian Manifolds
}

\author{%
	Tilman Daab, Noémie Jaquier, Christian Dreher, Andre Meixner, Franziska Krebs, and Tamim Asfour%
	\thanks{\ackHariaJubot}%
	\thanks{%
		The authors are with the Institute for Anthropomatics and Robotics, High Performance Humanoid Technologies Lab (H2T), at the Karlsruhe Institute of Technology (KIT), Karlsruhe, Germany.
		{\tt\small \{tilman.daab, asfour\}@kit.edu}%
	}
}

\begin{document}
 
\makeatletter
\let\@oldmaketitle\@maketitle
\renewcommand{\@maketitle}{\@oldmaketitle
	\vspace{-13ex}
}
\makeatother
\maketitle
\thispagestyle{empty}
\pagestyle{empty}

\captionsetup{font=footnotesize}

\begin{strip}%
    \centering%
    \captionsetup{type=figure}%
    \begin{subfigure}{0.14\textwidth}%
        \includegraphics[width=\linewidth]{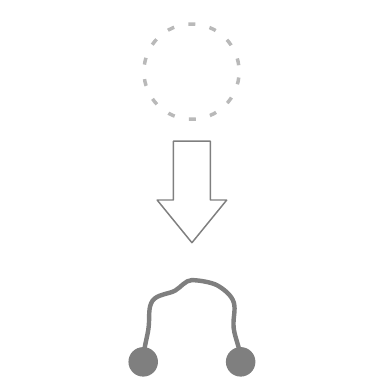}%
        \caption{\parbox[t]{4em}{Adding\linebreak}}%
        \label{fig:classificationIncrementalLearningOfMPs:add}%
    \end{subfigure}%
    \begin{subfigure}{0.14\textwidth}%
        \includegraphics[width=\linewidth]{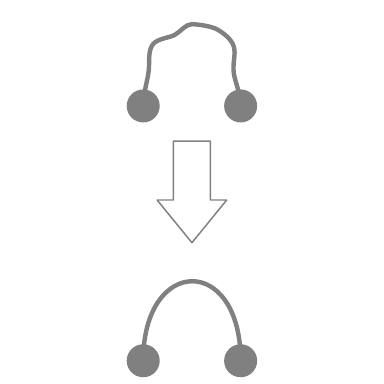}%
        \caption{\parbox[t]{4em}{Improving\linebreak}}%
        \label{fig:classificationIncrementalLearningOfMPs:improve}%
    \end{subfigure}%
    \begin{subfigure}{0.14\textwidth}%
        \includegraphics[width=\linewidth]{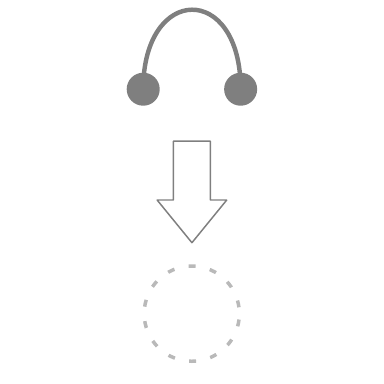}%
        \caption{\parbox[t]{4em}{Removing\linebreak}}%
        \label{fig:classificationIncrementalLearningOfMPs:remove}%
    \end{subfigure}%
    \begin{subfigure}{0.14\textwidth}%
        \includegraphics[width=\linewidth]{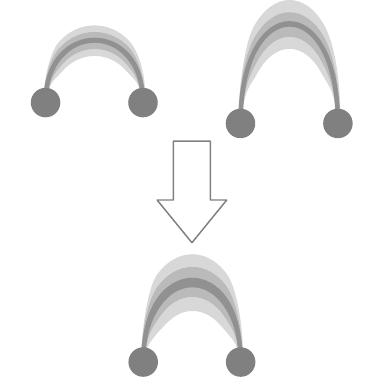}%
        \caption{\parbox[t]{5em}{Merging\linebreak modes}}%
        \label{fig:classificationIncrementalLearningOfMPs:mergeModes}%
    \end{subfigure}%
    \begin{subfigure}{0.14\textwidth}%
        \includegraphics[width=\linewidth]{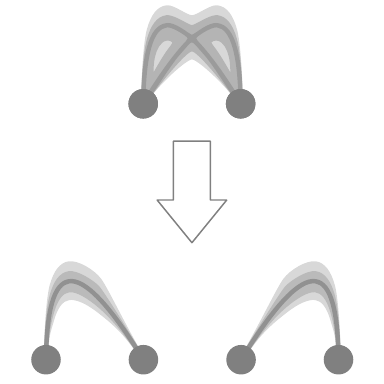}%
        \caption{\parbox[t]{5em}{Splitting\linebreak a mode}}%
        \label{fig:classificationIncrementalLearningOfMPs:splitMode}%
    \end{subfigure}%
    \begin{subfigure}{0.14\textwidth}%
        \includegraphics[width=\linewidth]{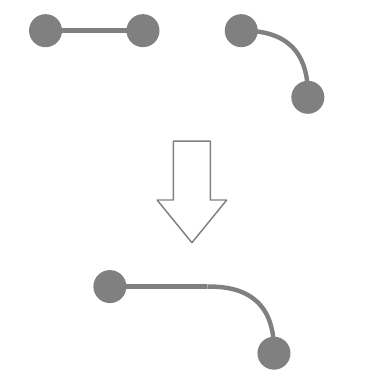}%
        \caption{\parbox[t]{5em}{Merging\linebreak temporally}}%
        \label{fig:classificationIncrementalLearningOfMPs:mergeTemporally}%
    \end{subfigure}%
    \begin{subfigure}{0.14\textwidth}%
        \includegraphics[width=\linewidth]{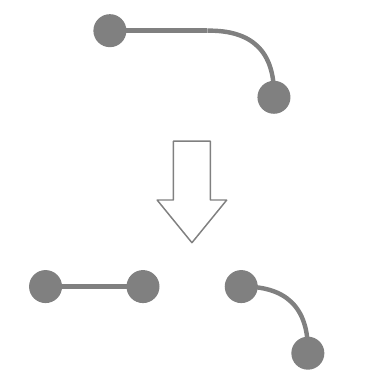}%
        \caption{\parbox[t]{5em}{Splitting\linebreak temporally}}%
        \label{fig:classificationIncrementalLearningOfMPs:splitTemporally}%
    \end{subfigure}%
    \caption{%
        Proposed fundamental operations to incrementally learn MP libraries.
        We provide formulations of the spatial operations (a)--(e) for VMPs.
    }%
    \label{fig:classificationIncrementalLearningOfMPs}%
\end{strip}

\begin{abstract}
    Movement primitives (MPs) are compact representations of robot skills that can be learned from demonstrations and combined into complex behaviors.
    However, merely equipping robots with a fixed set of innate MPs is insufficient to deploy them in dynamic and unpredictable environments.
    Instead, the full potential of MPs remains to be attained via adaptable, large-scale MP libraries.
    In this paper, we propose a set of seven fundamental operations to incrementally learn, improve, and re-organize MP libraries.
    To showcase their applicability, we provide explicit formulations of the spatial operations for libraries composed of Via-Point Movement Primitives (VMPs).
    By building on Riemannian manifold theory, our approach enables the incremental learning of all parameters of position and orientation VMPs within a library.
    Moreover, our approach stores a fixed number of parameters, thus complying with the essential principles of incremental learning.
    We evaluate our approach to incrementally learn a VMP library from motion capture data provided sequentially.
\end{abstract}

\section{Introduction}
\label{sec:introduction}

Robots providing help to humans at home, taking care of the elderly, or collaborating with workers must be able to learn new skills and to adapt them to unseen situations. 
In this context, learning from demonstrations~\cite{billard2008handbookOfRobotics} and imitation learning~\cite{schaal1999route} are promising frameworks to learn generalizable skills from human demonstrations. Such skills are often represented in the form of movement primitives (MPs)~\cite{Giszter1993,schaal1999route},
and may be stored to form a MP library~\cite{schaal1999route,pastor2009motorSkills}. Learning and adapting skills thus essentially entails incrementally updating the library.
Different definitions of incremental learning exist in the literature, see, e.g.,~\cite{GepperthHammer2016incremental,losingIncrementalOnlineLearning2018} for an overview.
Akin to Gepperth and Hammer~\cite{GepperthHammer2016incremental}, we consider
\emph{online learning} as learning from sequentially-arriving data, and \emph{incremental learning} as online learning with limited amount of memory.
Specifically, we require the memory for learning $m$ models from $n$ samples with $m < n$ to be bound by $\mathcal{O}(f(m))$, i.e., independently of~$n$.
This implies that approaches requiring to permanently store all samples or individually-retrieved sample representations are not considered as incremental.

In this paper, we tackle the problem of incrementally learning a MP library. Specifically, we argue that this goes beyond adding and improving existing MPs and introduce seven spatial and temporal operations to extend, update, and re-organize MP libraries based on incoming new demonstration examples (see Figure~\ref{fig:classificationIncrementalLearningOfMPs} and Section~\ref{sec:incremental_learning:classification_of_capabilities}). To showcase their applicability, we provide an explicit formulation of the spatial operations for a library of Via-Point Movement Primitives (VMPs)~\cite{zhou2019vmp} as MPs. Complex robotic skills also require to handle full-pose (i.e., positions and orientations) end-effector trajectories. Therefore, we provide methods to incrementally adapt all parameters of full-pose VMPs using the fundamental spatial operations (Section~\ref{sec:FullPoseVMP}). By building on Riemannian manifolds theory (Section~\ref{subsec:RiemannianVMP}), our approach soundly treats full-pose trajectories, while adhering to the aforementioned incremental learning definition.

The contributions of this paper are threefold:
\emph{(i)} We introduce a set of fundamental operations to incrementally learn MP libraries;
\emph{(ii)} We provide formulations to incrementally learn all parameters (\ie, weights, via-points, and task parameters) of the VMPs within a library according to the fundamental spatial operations;
and
\emph{(iii)} in doing so, we provide a Riemannian formulation of VMPs to handle full-pose trajectories.
We showcase our approach by incrementally learning a VMP library from human demonstrations recorded via motion capture.

\section{Related Work}

In the following, we review existing works in the field of incremental learning of MPs.
The influential works~\cite{kulic2012primitiveTree,niekum2013fsa,niekumLearningGroundedFiniteState2015} are examples of online learning of MPs.
Kulić \etal~\cite{kulic2012primitiveTree} learn MPs from dynamically-clustered demonstrations.
This allows adding, improving, and splitting modes of MPs. However, this approach is not incremental as it permanently stores a model of each demonstration.
Niekum \etal~\cite{niekum2013fsa,niekumLearningGroundedFiniteState2015}
learn dynamical movement primitives (DMPs)~\cite{ijspeert2013dmp} and represent tasks as graphs.
The model can be changed arbitrarily by re-creating it, but not via specific operations.
The approach is not incremental as all demonstrations are stored.
Gutierrez \etal~\cite{gutierrezIncrementalTaskModification2018,gutierrezLearningCorrectiveDemonstrations2019} proposed to add and improve MPs online based on HMMs.

Existing incremental approaches often only support a single operation to modify models in a MP library.
Adding was formulated in~\cite{meier2012regonition} for DMPs.
Improving was formulated for periodic DMPs~\cite{gamsAdaptationCoachingPeriodic2016}, for MPs based on Gaussian Mixture Models (GMMs)~\cite{calinonIncrementalLearningGestures2007}, for Dynamical Systems (DS)~\cite{kronanderIncrementalMotionLearning2015}, for HMMs~\cite{takano2016incrementalHMM}, and for expected sensor data during the execution of DMPs~\cite{pastorAssociativeSkillMemories2012}.
Adding, improving and removing as three operations were formulated for DMPs specifically in 2D positional space \cite{lemme2014bootstrapping}.
In this paper, we instead propose seven fundamental operations of incrementally learning a MP library and formulate the five spatial ones within a single framework.

\section{Background}

In this section, we introduce the MP formulation on which we focus in this paper, as well as the mathematical tools that are required to handle full-pose demonstrations.

\subsection{Via-Point Movement Primitives}
\label{subsec:BackgroundVMP}

Via-Point Movement Primitives (VMPs)~\cite{zhou2019vmp} are highly flexible MPs that combine the extrapolation capabilities of DMPs~\cite{ijspeert2013dmp} with the ability of probabilistic movement primitives (ProMPs)~\cite{paraschos2013promp} to handle via-points.
A VMP models a trajectory $\bm{y}(\varphi)$ as the superposition of an elementary trajectory $\bm{h}(\varphi)$ and a shape modulation $\bm{f}(\varphi)$, i.e.,
\begin{equation}
    \label{eqn:classical_vmp:overall_trajectory}
	\bm{y}(\varphi) = \bm{h}(\varphi) + \bm{f}(\varphi),
\end{equation}
where $\varphi\in[0, 1]$ denotes the phase variable allowing for temporal modulation. 
The elementary trajectory forms the basic structure of the VMP by connecting two consecutive points such as a start $\bm{h}_0$ and a goal $\bm{h}_1$. Without loss of generality, we consider linear elementary trajectories
\begin{equation}
    \label{eqn:classical_vmp:elementary_trajectory:basic_case_without_via_points}
	\bm{h}(\varphi) = \bm{h}_0 + \varphi \cdot (\bm{h}_1 - \bm{h}_0).
\end{equation}
The shape modulation is represented as a weighted sum of basis functions, so that 
\begin{equation}
    \label{eqn:classical_vmp:shape_modulation}
	\bm{f}(\varphi) =
    \Psi(\varphi)
    \bm{w},
\end{equation}
where $\Psi(\varphi) \in \realR^{d\times dk}$ is an activation matrix computed as a block-diagonal matrix of radial basis functions (RBFs) $\bm{\psi}(\varphi)^T \in \realR^{1\times k}$, and $\bm{w} = [\bm{w}_1^T, \dots, \bm{w}_k^T]^T \in \realR^{dk}$ is a concatenated weight vector.
As in ProMPs, $\bm{w}$ follows a Gaussian distribution $\mathcal{N}(\bm{\mu}_{\bm{w}}, \bm{\Sigma}_{\bm{w}})$, whose parameters are computed from the weight vectors estimated for each demonstration.
After the initial learning phase, $\bm{f}(\varphi)$ can be adapted via the conditioning of $\bm{w}$.
Alternatively, and as preferred by Zhou \etal~\cite{zhou2019vmp} for orientations, adaptations can be achieved by modulating the elementary trajectory using a via-point.

Given a desired pose $\bm{y}_{v} = \bm{y}(\varphi_{v})$ associated with a phase value $\varphi_{v}$, the pose of an elementary via-point is
\begin{equation}
    \label{eqn:classical_vmp:via-point}
    \bm{h}_v = \bm{y}_v - \bm{f}(\varphi_v).
\end{equation}
With $v^-$ and $v^+$ denoting the surrounding via-points of a phase, the elementary trajectory~\eqref{eqn:classical_vmp:elementary_trajectory:basic_case_without_via_points} becomes
\begin{equation*}
	\bm{h}(\varphi) = \bm{h}_{v^-} + \frac{\varphi - \varphi_{v^-}}{\varphi_{v^+} - \varphi_{v^-}} \cdot (\bm{h}_{v^+} - \bm{h}_{v^-}).
\end{equation*}
Note that the start and end of the VMP are simply considered as via-points for $\varphi=0$ and $\varphi=1$, respectively.

Since quaternions must satisfy a unit-norm constraint, additional care must be taken to formulate orientation VMPs. Zhou \etal~\cite{zhou2019vmp} defined the elementary trajectory via spherical linear interpolation.
Trajectories $\bm{y}(\varphi)$ were then modeled via quaternion multiplication of $\bm{h}(\varphi)$ with a shape modulation $\bm{f}(\varphi)$.
The vector part of $\bm{f}(\varphi)$ was learned using~\eqref{eqn:classical_vmp:shape_modulation}, and its scalar part deduced to respect the unit-norm constraint.
However, such normalization operations are known to lead to inaccurate models~\cite{rozoOrientationProbabilisticMovement2021,Zhang23:RiemannianStableDS}.
In this paper, we instead account for the intrinsic geometry of quaternions and formulate a Riemannian approach to incrementally learn full-pose VMPs.

\subsection{Riemannian Manifolds}
In this paper, we represent orientations via unit quaternions, which are widely-used nearly-minimal representations. As quaternions must satisfy a unit-norm constraint, they cannot be treated with traditional Euclidean methods. Instead, we leverage tools from Riemannian geometry to learn orientation trajectories. A smooth manifold $\manifoldM$ is a curved topological space endowed with a smooth differential structure~\cite{Lee13:SmoothManifolds}. A tangent space $\tangent{\manifoldM}{\bm{x}}$, i.e., a Euclidean approximation of the manifold, is associated with each point $\bm{x}\in\manifoldM$. A Riemannian manifold is a smooth manifold equipped with a Riemannian metric, i.e., a smoothly-varying positive-definite inner product acting on each tangent space $\tangent{\manifoldM}{\bm{x}}$~\cite{Lee18:RiemannianManifolds}. The Riemannian metric defines geodesics as shortest paths between two points on $\manifoldM$, thus generalizing the notion of straight lines to Riemannian manifolds. Euclidean tangent spaces can be leveraged via the use of the exponential map $\Exp{\bm{x}}{\bm{u}}: \tangent{\manifoldM}{\bm{x}} \to \manifoldM$ and logarithmic map $\Log{\bm{x}}{\bm{y}}: \manifoldM \to \tangent{\manifoldM}{\bm{x}}$. Finally, the parallel transport operates $\parallelTransport{\bm{x}}{\bm{y}}{\bm{u}}:\tangent{\manifoldM}{\bm{x}} \to\tangent{\manifoldM}{\bm{y}}$ with tangent vectors lying on different tangent spaces. Full-pose trajectories are composed of points on the product of manifold $\manifoldM=\realR^3\times \manifoldS^3$, where positions and quaternions are elements of the Euclidean space $\realR^3$ and sphere manifold $\manifoldS^3$, respectively.

\section{Operations to incrementally learn Movement Primitive Libraries}
\label{sec:incremental_learning:classification_of_capabilities}

As stated in \Cref{sec:introduction}, learning a movement primitive library incrementally goes beyond adding or incrementally improving single MPs.
Here, we propose to structure the incremental learning capabilities of MP libraries into seven fundamental operations (see Figure~\ref{fig:classificationIncrementalLearningOfMPs}).
These operations not only incorporate new knowledge but also provide the ability to revert operations that were taken erroneously. This is crucial, as systems that decide which operation to perform hardly work perfectly. The three first operations are:
\begin{enumerate}[(a)]
    \item \emph{Adding} a MP to the library to extend it. This is needed if a new observation does not match any existing MP in the library, \eg, as in~\cite{meier2012regonition}.
    \item \emph{Improving} an existing MP. This allows generalization based on multiple demonstrations, \eg, as in~\cite{gamsAdaptationCoachingPeriodic2016}.
    \item \emph{Removing} a MP that is not required anymore, \eg, as in~\cite{lemme2014bootstrapping}, or was added erroneously,
    e.g., as a consequence of perception problems.
\end{enumerate}
In the following, we refer to multiple distinct ways to perform a movement as \emph{modes}.
The last four operations describe ways of re-organizing a MP library, namely:
\begin{enumerate}[(a)]
\setcounter{enumi}{3} 
    \item \emph{Merging modes} to merge two separately-stored MPs representing similar movements into a single one.
    \item \emph{Splitting a mode} into two distinct ones, each of them represented by a single MP, \eg, as (online) in~\cite{kulic2012primitiveTree}. This is useful when a MP consists of several modes that were erroneously learned as a single one.
    \item \emph{Merging temporally} several MPs that always occur together or correspond to a movement that was over-segmented into multiple MPs.
    \item \emph{Splitting temporally} a longer MP which, inversely, consists of several movements that could be used separately.
\end{enumerate}
Given these fundamental operations, key challenges are how, when, and which of them to apply within MP libraries.
We start by tackling the first of these challenges and formulate the five spatial operations (a)--(e) in a specific MP library.

\section{Incremental Learning of Full-Pose Via-Point Movement Primitives}
\label{sec:FullPoseVMP}

In this section, we specifically address the problem of incrementally learning a library of Via-Point Movement Primitives (VMPs)~\cite{zhou2019vmp}. As many robot skills are composed of position and orientation, we first introduce a Riemannian formulation for full-pose VMPs representing trajectories $\bm{y}\in\realR^3 \times \manifoldS^3$. Then, we provide methods to incrementally learn full-pose VMPs within our library via the $5$ spatial operations (a)--(e) introduced in the previous section.
In this paper, we assume that demonstrations are provided with a perfect segmentation.
Thus, temporal operations will be addressed as future work.

\subsection{Full-Pose VMPs}
\label{subsec:RiemannianVMP}
Defining $\manifoldM = \realR^3 \times \manifoldS^3$, the VMP basic principle translates to the Riemannian case as follows.
The trajectory $\bm{y}(\varphi)\in\manifoldM$ is composed by an elementary trajectory $\bm{h}(\varphi)\in\manifoldM$, which is modified by a shape modulation $\bm{f}_{\bm{h}(\varphi)}(\varphi)\in\tangent{\manifoldM}{\bm{h}(\varphi)}$ as
\begin{equation}
    \label{eq:pose_vmp:overall}
    \bm{y}(\varphi) = \Exp{\bm{h}(\varphi)}{\bm{f}_{\bm{h}(\varphi)}(\varphi)},
\end{equation}
with phase variable $\varphi\in[0,1]$.
The elementary trajectory connects a start $\bm{h}_0$ to a goal $\bm{h}_1$ via a geodesic
\begin{equation}
    \label{eqn:elementary_trajectory:basic_case_without_via_points}
    \bm{h}(\varphi) = \Exp{\bm{h}_0}{ \varphi \cdot \Log{\bm{h}_0}{\bm{h}_1} }.
\end{equation}
Note that Eqs.~\eqref{eq:pose_vmp:overall} and~\eqref{eqn:elementary_trajectory:basic_case_without_via_points} simplify to~\eqref{eqn:classical_vmp:overall_trajectory} and~\eqref{eqn:classical_vmp:elementary_trajectory:basic_case_without_via_points}, respectively, when $\manifoldM=\realR^d$, i.e., we recover the Euclidean VMP.

Importantly, the shape modulation is an element of the tangent space at the elementary trajectory. Thus, it follows a Euclidean structure and can be defined similarly as in~\eqref{eqn:classical_vmp:shape_modulation} with special care of defining common weight vectors across the different tangent spaces $\tangent{\manifoldM}{\bm{h}(\varphi)}$.
To do so, we define the weights as $\bm{w}_{\bm{h}_{0}}\in\tangent{\manifoldM}{\bm{h}_0}$ and obtain $\bm{f}_{\bm{h}(\varphi)}$ as
\begin{equation}
    \label{eq:shape_modulation_at_point_of_elementary_trajectory}
    \bm{f}_{\bm{h}(\varphi)}(\varphi) = \parallelTransport{\bm{h}_{0}}{\bm{h}(\varphi)}{\bm{f}_{\bm{h}_{0}}(\varphi)
    },
\end{equation}
with $\bm{f}_{\bm{h}_{0}}(\varphi) = \Psi(\varphi) \bm{w}_{\bm{h}_{0}}$.
Note that parallel transport conserves the angle between shape modulation and the direction of the elementary trajectory.

\begin{figure}
    \centering
    \begin{subfigure}[b]{.5\linewidth}
        \centering
        \begin{tikzpicture}[inner xsep=0pt, inner ysep=0pt, outer xsep=0pt, outer ysep=0pt]
            \node [anchor=south west] (image) at (0, 0) {
                \includegraphics[height=0.65\linewidth]{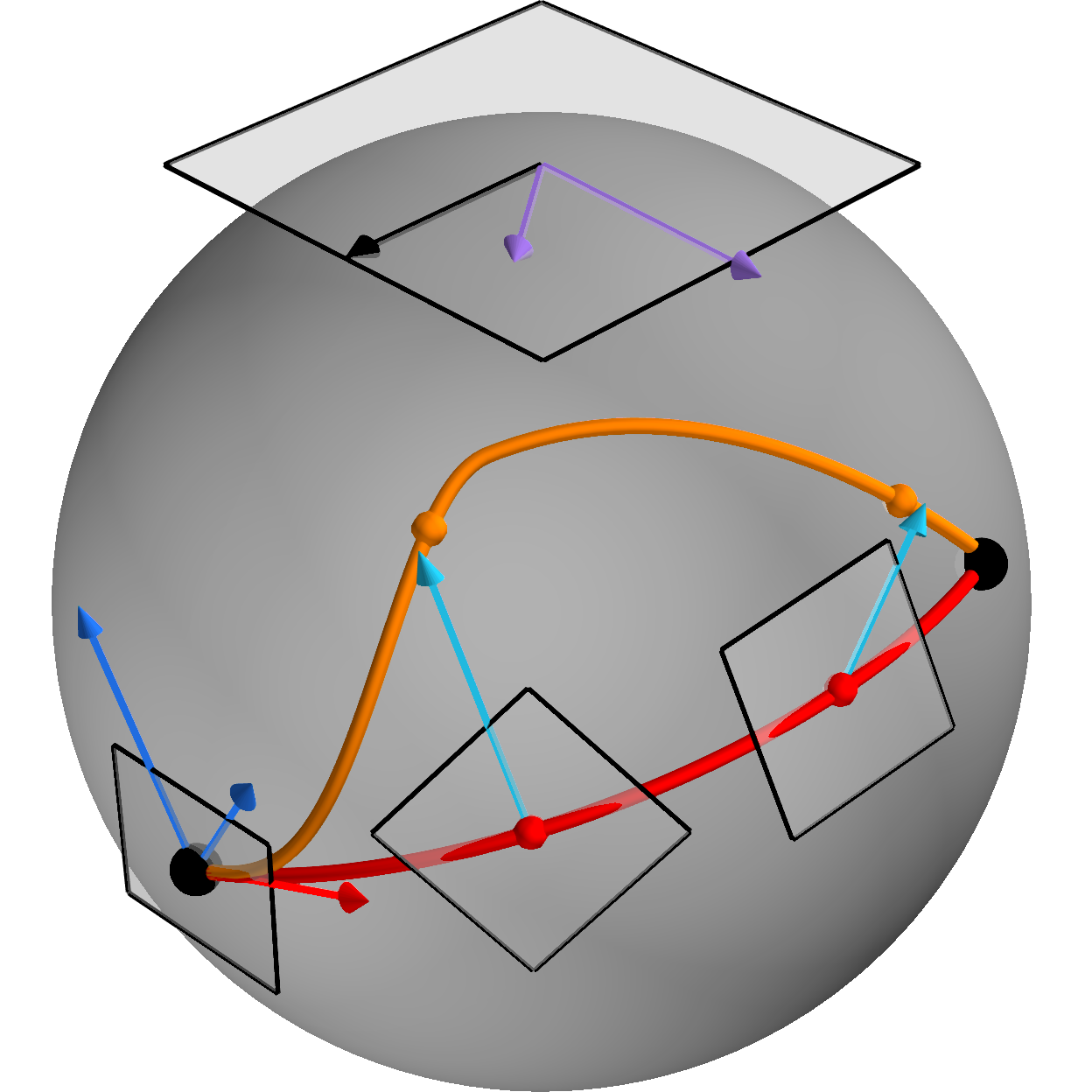}
            };
            \begin{scope}[
                x={($(image.south east)$)},
                y={($(image.north west)$)}
            ]

\draw (0.2,  0.97) node [left=0.2em] {\footnotesize$\bm{d}_{\bm{o}}$} --
      (0.36, 0.81)      [line width=0.75pt, black];
\draw (0.1 , 0.71) node [left=0.2em] {\footnotesize$\bm{f}_{\bm{h}(\varphi)}(\varphi)$} --
      (0.42,  0.35)      [line width=0.75pt, black];
\draw (0,    0.41) node [left=0.2em] {\footnotesize$\bm{f}_{\bm{h}_0}(\varphi)$} --
      (0.1 , 0.35)      [line width=0.75pt, black];
\draw (0.05,    0.22) node [left=0.2em] {\footnotesize$\bm{h}_{0}$} --
      (0.145, 0.2)      [line width=0.75pt, black];
\draw (0.2,    0.03) node [left=0.2em] {\footnotesize$\bm{d}_{01}$} --
      (0.3, 0.16)      [line width=0.75pt, black];

\draw (0.8,  0.97) node [right=0.2em] {\footnotesize$\bm{f}_{\bm{o}}(\varphi)$} --
      (0.63, 0.81)      [line width=0.75pt, black];
\draw (0.9,  0.78) node [right=0.2em] {\footnotesize$\bm{y}(\varphi)$} --
      (0.68, 0.63)      [line width=0.75pt, black];
\draw (0.95, 0.59) node [right=0.2em] {\footnotesize$\bm{h}_1$} --
      (0.92, 0.52)      [line width=0.75pt, black];
\draw (0.9,  0.22) node [right=0.2em] {\footnotesize$\bm{h}(\varphi)$} --
      (0.66, 0.27)      [line width=0.75pt, black];
                \end{scope}
			\end{tikzpicture}
        \caption{Shape modulation transport}
        \label{fig:RiemannianVMP:IllustrationS2:shapeModulationTransport}%
    \end{subfigure}%
    \begin{subfigure}[b]{.5\linewidth}
        \centering
        \begin{tikzpicture}[inner xsep=0pt, inner ysep=0pt, outer xsep=0pt, outer ysep=0pt]
            \usetikzlibrary{calc}
            \node [anchor=south west] (image) at (0, 0) {
                \includegraphics[width=0.65\linewidth]{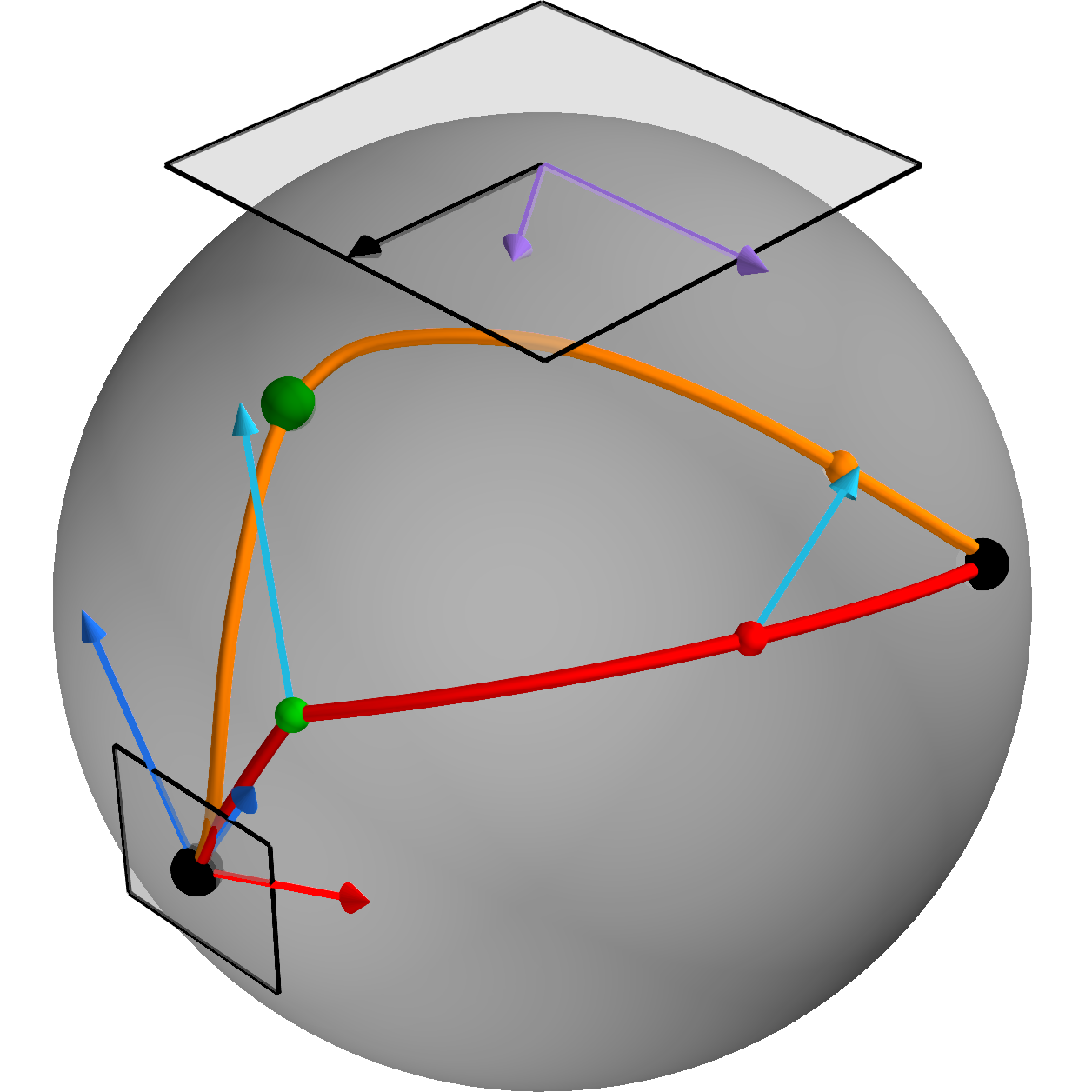}
            };
            \begin{scope}[
                x={($(image.south east)$)},
                y={($(image.north west)$)}
            ]

\draw (0.1,  0.78) node [left=0.2em] {\footnotesize$\bm{y}(\varphi_v)$} --
      (0.24, 0.66)      [line width=0.75pt, black];
\draw (0.04, 0.59) node [left=0.2em] {\footnotesize$\bm{f}_{\bm{h}(\varphi)}(\varphi)$} --
      (0.24, 0.42)      [line width=0.75pt, black];
\draw (0,    0.03) node [left=0.2em] {\footnotesize$\bm{d}_{01}$} --
      (0.29, 0.165)      [line width=0.75pt, black];

\draw (0.9,  0.78) node [right=0.2em] {\footnotesize$\bm{y}(\varphi)$} --
      (0.7 , 0.64)      [line width=0.75pt, black];
\draw (0.9,  0.22) node [right=0.2em] {\footnotesize$\bm{h}(\varphi)$} --
      (0.55, 0.36)      [line width=0.75pt, black];
\draw (0.8,  0.03) node [right=0.2em] {\footnotesize$\bm{h}(\varphi_v)$} --
      (0.29, 0.32)      [line width=0.75pt, black];
				\end{scope}
			\end{tikzpicture}
        \caption{Execution with via-point}
        \label{fig:RiemannianVMP:IllustrationS2:ExecutionWithViaPoint}%
    \end{subfigure}
    \caption{Illustration of a Riemannian VMP on $\manifoldS^2$.}
    \label{fig:RiemannianVMP:IllustrationS2}
    \vspace{-0.5cm}
\end{figure}

Demonstrations characterized by different weight vectors $\bm{w}_{\bm{h}_{0}, n}$ lead to a distribution $\bm{w}\sim\mathcal{N}(\bm{\mu}_{\bm{w}}, \bm{\Sigma}_{\bm{w}})$.
The computation of this distribution is hindered by the fact that each $\bm{w}_{\bm{h}_{0},n}$ lies on a different tangent space $\tangent{\manifoldM}{\bm{h}_{0,n}}$. 
This is resolved by transporting all $\bm{f}_{\bm{h}_{0}}$ to a common tangent space $\tangent{\manifoldM}{\bm{o}}$ defined, e.g., at the origin $\bm{o}$ of $\manifoldM$.
However, in this case, parallel transport does not suffice to ensure consistency between shape modulation vectors in $\tangent{\manifoldM}{\bm{o}}$. 
Specifically, two identical trajectories translated on the manifold would be encoded in $\tangent{\manifoldM}{\bm{o}}$ by shape modulation vectors differing by a rotation depending on their starting points. 
We compensate this by aligning the parallel-transported direction $\bm{d}_{01} = \Log{\bm{h}_{0}}{\bm{h}_{1}}/\|\Log{\bm{h}_{0}}{\bm{h}_{1}}\|_{\bm{h}_{0}} \in \tangent{\manifoldM}{\bm{h}_{0}}$ of each demonstration with a default unit direction $\bm{d}_{\bm{o}} \in \tangent{\manifoldM}{\bm{o}}$.
This is achieved via a rotation $\bm{R}$ satisfying
\begin{equation}
    \label{eqn:definitionOfRotation}
    \bm{d}_{\bm{o}} = \bm{R} \,\parallelTransport{\bm{h}_{0}}{\bm{o}}{\bm{d}_{01}}.
\end{equation}
The shape modulation vectors are then obtained as 
\begin{equation*}
    \bm{f}_{\bm{o}} = \bm{R} \, \parallelTransport{\bm{h}_{0}}{\bm{o}}{
        \bm{f}_{\bm{h}_{0}}
    }.
\end{equation*}
Figure \ref{fig:RiemannianVMP:IllustrationS2:shapeModulationTransport} illustrates the transport for two shape modulations.
If $\bm{h}(0)=\bm{h}(1)$, $\bm{d}_{01}$ is ill-defined, and the rotation is omitted.
Notice that, in general, $\bm{y}_0\approx \bm{h}_0$ and $\bm{y}_1\approx \bm{h}_1$ hold aside from approximation errors of $\bm{f}(0)$ and $\bm{f}(1)$.

Similarly as Euclidean VMPs, adaptation of Riemannian VMPs can be achieved by conditioning $\bm{w}$ or by modulating the elementary trajectory with via-points.
Analougously to Eq.~\eqref{eqn:classical_vmp:via-point} in Euclidean spaces,
elementary via-points $\bm{h}_{v}$ at $\varphi_{v}$ are derived from trajectory via-points $\bm{y}_{v}=\bm{y}(\varphi_v)$ so that
\begin{equation}
    \label{eq:via_point}
    \bm{y}_{v} = \Exp{\bm{h}_{v}}{\bm{f}_{\bm{h}_v}(\varphi_{v})} = \Exp{\bm{h}_{v}}{
        \parallelTransport{\bm{h}_{0}}{\bm{h}_{v}}{\bm{f}_{\bm{h}_{0}}(\varphi_{v})}
    }.
\end{equation}
In contrast to the Euclidean case, Eq.~\eqref{eq:via_point} does not yield an analytical solution. Instead, it resembles a geodesic regression problem~\cite{fletcher2013geodesicRegression} and a solution $\bm{h}_v$ can be calculated by minimizing
the distance $d^2_{\manifoldM}(\bm{y}_{v}, \tilde{\bm{y}}_{v})$ between the desired pose $\tilde{\bm{y}}_{v}$ and the pose~\eqref{eq:via_point}. This is achieved via the Riemannian gradient descent update
\begin{equation*}
   \bm{h}_{v,i+1} \gets \Exp{\bm{h}_{v,i}}{
        2 \alpha
        \parallelTransport{\bm{y}_{v,i}}{\bm{h}_{v,i}}{
            \Log{\bm{y}_{v,i}}{\tilde{\bm{y}}_v}
        }
    },
\end{equation*}
where we use a gradient approximation akin to and the adaptive stepsize $\alpha$ of~\cite{kim2014multivariateLinearRegression}.
Finally, Eq.~\eqref{eqn:elementary_trajectory:basic_case_without_via_points} becomes
\begin{equation*}
    \bm{h}(\varphi) = \Exp{\bm{h}_{v^-}}{ \frac{\varphi - \varphi_{v^-}}{\varphi_{v^+} - \varphi_{v^-}} \cdot \Log{\bm{h}_{v^-}}{\bm{h}_{v^+}} }.
\end{equation*}
Note that $\bm{R}$ is still defined as a function of the direction $\bm{d}_{01}$ to avoid discontinuities in $\bm{y}(\varphi)$ in the presence of via-points.
Figure~\ref{fig:RiemannianVMP:IllustrationS2:ExecutionWithViaPoint} illustrates the execution of a VMP with a via-point.
Next, we show how to apply the fundamental spatial operations of Section~\ref{sec:incremental_learning:classification_of_capabilities} to incrementally learn the weights, via-points, and task parameters of full-pose VMPs.

\subsection{Incremental Learning of VMP Weights}
\label{subsec:incremental_learning:vmp_weights}
Following the definition of Section~\ref{sec:introduction}, incremental learning of full-pose VMPs is achieved by storing only a fixed amount of parameters while demonstrations are provided sequentially.
In this section, we present incremental approaches to learn the weights $\bm{w}$ characterizing incrementally-learned VMPs.
The building blocks $\bm{w}_i$ of the concatenated weight vector $\bm{w}$ lie in tangent spaces of $\realR^3 \times\manifoldS^3$ and thus display a Euclidean structure.
Therefore, $\bm{w}$ is learned incrementally as the Euclidean distribution $\bm{w}\sim\mathcal{N}(\bm{\mu}_{\bm{w}}, \bm{\Sigma}_{\bm{w}})\subset\realR^{7k}$.

Given $N$ samples $\{\bm{x}_n\}_{n=1}^N$, a Gaussian distribution $\mathcal{N}(\bm\mu, \bm{\Sigma}) \in \realR^d$ is estimated batch-wise following
\begin{equation}
	\hat{\bm{\mu}}_n = \frac{1}{n} \sum_{i=1}^{n} \bm{x}_i \;\;\text{and}\;\;
	\label{eqn:batchwise_scalar_variance_estimation}
	\hat{\bm{\Sigma}}_n = \frac{1}{n-1} \sum_{i=1}^{n} (\bm{x}_i - \bm{\hat\mu}_n) (\bm{x}_i - \bm{\hat\mu}_n)^\trsp,
\end{equation}
with $\hat{\bm{\Sigma}}_n$ being defined if $n\geq2$.
Internally, we represent the estimated Gaussian distribution with three parameters: The number of samples $n$ and the two estimated non-central moments $\hat{\bm{\mu}}_n = \expectation{X} $ and $\hat{\bm{S}}_n = \expectation{XX^\trsp}$. Thus, we have
\begin{equation}
    \label{eqn:covariance}
    \hat{\bm{\Sigma}}_n = \frac{n}{n-1} \left( \hat{\bm{S}}_n - \hat{\bm{\mu}}_n \hat{\bm{\mu}}_n^\trsp \right). 
\end{equation}
The weight distribution parameters are incrementally learned to achieve the five spatial operations of Section~\ref{sec:incremental_learning:classification_of_capabilities} as follows.
\begin{enumerate}[(a)]
\item \emph{Adding} a VMP to the library is realized by creating a weight estimator. Based on a new demonstration $\bm{x}_1$ it is initialized with $n=1$, $\hat{\bm{\mu}}_1 = \bm{x}_1$ and $\bm{\hat{S}}_1 = \bm{x}_1 \bm{x}_1^\trsp$.

\item \emph{Improving} a weight estimation based on an additional demonstration $\bm{x}_{n+1}$ is achieved as
\begin{equation*}
	\hat{\bm{\mu}}_{n+1} = \frac{
        n \hat{\bm{\mu}}_{n} + \bm{x}_{n+1}
    }{n+1}
     \;\;\text{and}\;\;
	\bm{\hat{S}}_{n+1} = \frac{
        n \bm{\hat{S}}_{n} + \bm{x}_{n+1} \bm{x}_{n+1}^\trsp
    }{n+1}.
\end{equation*}

\item \emph{Removing} a VMP simply corresponds to deleting its weight estimator and potential references in task models.

\item \emph{Merging two modes} $\mathsf{A}$ and $\mathsf{B}$ results in a joint mode $\mathsf{C}$, whose weight parameters are, with $n_\mathsf{C} = n_\mathsf{A} + n_\mathsf{B}$,
\begin{equation}
    \label{eqn:merge_modes}
	\hat{\bm{\mu}}_{\mathsf{C}} = \frac{n_\mathsf{A}}{n_\mathsf{C}} \hat{\bm{\mu}}_{\mathsf{A}} + \frac{n_\mathsf{B}}{n_\mathsf{C}} \hat{\bm{\mu}}_{\mathsf{B}} \,\;\text{and}\;\,
	\bm{\hat{S}}_{\mathsf{C}} = \frac{n_\mathsf{A}}{n_\mathsf{C}} \bm{\hat{S}}_{\mathsf{A}} + \frac{n_\mathsf{B}}{n_\mathsf{C}} \bm{\hat{S}}_{\mathsf{B}}.
\end{equation}

\item \emph{Splitting a mode} $\mathsf{C}$ into two modes $\mathsf{A}$ and $\mathsf{B}$ using a demonstration $\bm{x}$ requires additional assumptions, as there are more variables than constraints.
We assume \emph{(i)} the demonstration to be similar to one of the modes,
$\bm{\hat{\mu}}_{\mathsf{A}} = \bm{x}$, and \emph{(ii)} both modes to have contributed equally to the existing estimation, i.e., ${n_\mathsf{A} = n_\mathsf{B} = \frac{n_\mathsf{C}}{2}}$.
Then, Eq.~\eqref{eqn:merge_modes} leads to $\hat{\bm{\mu}}_{\mathsf{B}} = 2 \hat{\bm{\mu}}_{\mathsf{C}} - \hat{\bm{\mu}}_{\mathsf{A}}$.
While it might be tempting to fix $\bm{\hat{S}}_{\mathsf{A}} = \bm{x} \bm{x}^\trsp$ and derive $\bm{\hat{S}}_\mathsf{B}$ using Eq.~\eqref{eqn:merge_modes}, this always leads to $\hat{\bm{\Sigma}}_{\mathsf{A}}$ being degenerated, and can even lead to $\hat{\bm{\Sigma}}_{\mathsf{B}}$ not being positive semi-definite.
To avoid this, we further assume $\hat{\bm{\Sigma}}_{\mathsf{A}}=\hat{\bm{\Sigma}}_{\mathsf{B}} = \sigma^2 \bm{I}$ with $\sigma=\frac{1}{3} \| \hat{\bm{\mu}}_{\mathsf{A}} - \hat{\bm{\mu}}_{\mathsf{B}} \| $.
$\bm{\hat{S}}_{\mathsf{A}}$ and $\bm{\hat{S}}_{\mathsf{B}}$ follow using Eq.~\eqref{eqn:covariance}.
To speed up convergence during subsequent improvements, we finally reduce $n$ for each mode by a factor of $\frac{1}{2}$.
\end{enumerate}
Importantly, mathematical equivalence with batch-wise estimations holds for all operations except \emph{splitting a mode}. 
As such, they offer the benefits of incremental learning without introducing drawbacks compared to batch-wise learning.

\subsection{Detection of Via-Points}
\label{subsec:detectionOfViaPoints}

Demonstrations with undetected via-points lead to distorted weight estimations as via-points distort the trajectory.
As a counteraction, we tackle the detection of via-points in this section.
Their incremental generalization is addressed as a part of the task parameter estimation in Section~\ref{subsec:incremental_learning:task_parameters}.
An experimental comparison of the approaches follows in Section~\ref{subsec:experiments:viaPoints}.
Here, we denote the reproduction of a demonstration
$\{\bm{\tilde{y}}_n\}_{n=1}^{N}$
with a VMP as $\{\bm{y}(\varphi_n) = \bm{y}_n\}_{n=1}^{N}$, with $\bm{y}(0) = \bm{\tilde{y}}_1$ and $\bm{y}(1) = \bm{\tilde{y}}_N$.
Their poses differ by
\begin{equation*}
    d(\bm{\tilde{y}}_n,\bm{y}_n) = d_{\realR^3}(\bm{\tilde{y}}_n^{\bm{p}},\bm{y}_n^{\bm{p}}) + \alpha \, d_{\manifoldS^3}(\bm{\tilde{y}}_n^{\bm{q}},\bm{y}_n^{\bm{q}}),
\end{equation*}
with positions $\bm{y}^{\bm{p}}$, orientations $\bm{y}^{\bm{q}}$, and a weighting factor $\alpha$.
This difference can be caused by a need to improve the VMP's weights and by undetected via-points.
To reduce it, via-points are added until a termination criterion is fulfilled, e.g., $d(\bm{\tilde{y}}_n,\bm{y}_n)<\theta$ with a given threshold $\theta$.
Each via-point is created based on its phase variable $\varphi_v$ and the associated pose $\bm{\tilde{y}}_n$.
We propose three ways to select $\varphi_v$.
First, $\varphi_v$ can be chosen as the phase of the \emph{maximum distance}, i.e.,
\begin{equation*}
    \varphi_v = \argmax_{\varphi_n
    } d(\bm{\tilde{y}}(\varphi_n), \bm{y}(\varphi_n)).
\end{equation*}
Second,
a \emph{brute force} search can be conducted among all phase values of the demonstration.
Evaluating the VMP with a via-point at $\varphi_c$ yields a trajectory $\{\bm{y}^c_n\}_{n=1}^N$.
The value $\varphi_c$ leading to the lowest average distance is picked as
\begin{equation*}
    \varphi_v = \argmin_{\varphi_c
    } \frac{1}{N} \sum_n d(\bm{\tilde{y}}_n,\bm{y}^{c}_n).
\end{equation*}
Third,
$\varphi_v$ can be selected via a \emph{weighted phase average} as
\begin{equation*}
    \varphi_v = \frac{
        \sum_{n=1}^N d(\bm{\tilde{y}}(\varphi_n), \bm{y}(\varphi_n)) \, \varphi_n
    }{
        \sum_{n=1}^N d(\bm{\tilde{y}}(\varphi_n), \bm{y}(\varphi_n))
    }.
\end{equation*}
For all approaches, redundant and low-influence via-points can be removed at the end of the greedy selection phase.

\subsection{Incremental Learning of Task Parameters}
\label{subsec:incremental_learning:task_parameters}

While weights are sufficient to characterize a VMP, its execution requires additional task parameters, i.e., a start pose $\bm{y}_0$, an end pose $\bm{y}_1$, optional via-points $\{(\varphi_{v,n}, \bm{y}_{v,n})\}_{n=1}^N$, and a duration $t$.
Estimating $\bm{y}_0$, $\bm{y}_1$ and $\bm{y}_{v,n}\in\realR^3 \times \manifoldS^3$
amounts to estimating a position $\bm{p}\in \realR^3$ and orientation $\bm{q}\in \manifoldS^3$.
Moreover, $\varphi_{v,n}$ and $t$ are estimated as scalars.

Incremental estimation of scalars and positions is achieved as detailed in Section~\ref{subsec:incremental_learning:vmp_weights}.
Although we require $\varphi_{v,n} \in [0;1]$ and $t > 0$, we deem the Gaussian estimation to nevertheless be appropriate as we assume the variances to be low.
The batchwise mean on Riemannian manifolds is generalized via the so-called Fréchet mean~\cite{pennec2006basicTools}. When samples are provided sequentially, the orientation part of task parameters is obtained using the incremental Fréchet mean estimator~\cite{salehian2015ifme} as
\begin{equation}
    \bm{\hat{q}}_{n+1} = \Exp{\bm{\hat{q}}_n}{
        \frac{1}{n+1} \Log{\bm{\hat{q}}_n}{\bm{x}_{n+1}}
    }.
    \label{Eq:IncrementalFrechetMean}
\end{equation}
Although strict equivalence between batch-wise and incremental estimates does not apply in the Riemannian case, the estimator~\eqref{Eq:IncrementalFrechetMean} provably converges to the batch-wise Fréchet mean~\cite{salehian2015ifme}. This convergence is sufficient for our use case. Covariances are defined on tangent spaces $\tangent{\manifoldM}{\hat{\bm{q}}_n}$ as 
\begin{equation*}
\hat{\bm{\Sigma}}_n = \frac{1}{n-1} \sum_{i=1}^{n} \Log{\bm{\hat q}_n}{\bm{x}_i} \Log{\bm{\hat q}_n}{\bm{x}_i}^\trsp,
\end{equation*}
and incrementally learned as in Section~\ref{subsec:incremental_learning:vmp_weights} with special care of transporting the covariances to appropriate tangent spaces.
Finally, an executable task is represented by a task model containing a sequence of VMPs and their task parameters.

\section{Experiments}
\label{sec:experiments}
\begin{figure}
    \centering%
    \begin{subfigure}{0.2268\columnwidth}%
        \centering
        \includegraphics[height=2.75cm]{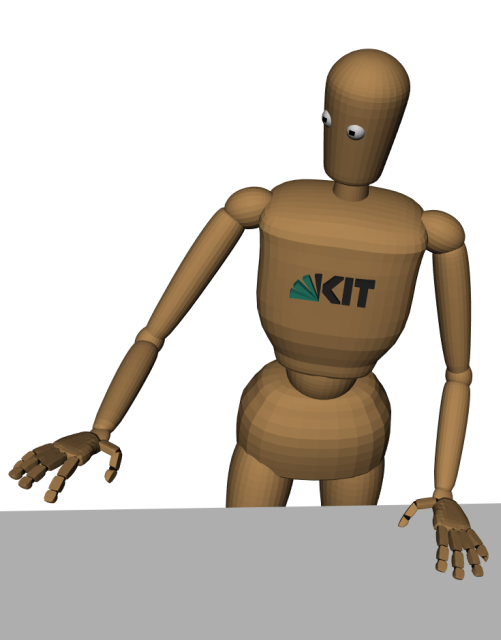}
        \caption{Approach}%
        \label{fig:experiments:mmm_recording_images:approach}%
    \end{subfigure}%
    \hfill%
    \begin{subfigure}{0.1854\columnwidth}%
        \centering
        \includegraphics[height=2.75cm]{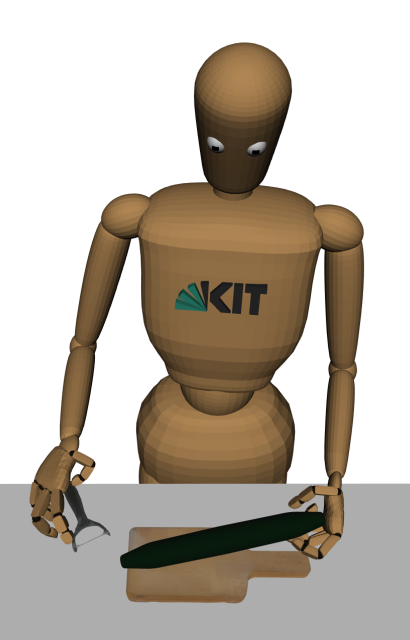}
        \caption{Lift}%
        \label{fig:experiments:mmm_recording_images:lift}%
    \end{subfigure}%
    \hfill%
    \begin{subfigure}{0.2675\columnwidth}%
        \centering
        \includegraphics[height=2.75cm]{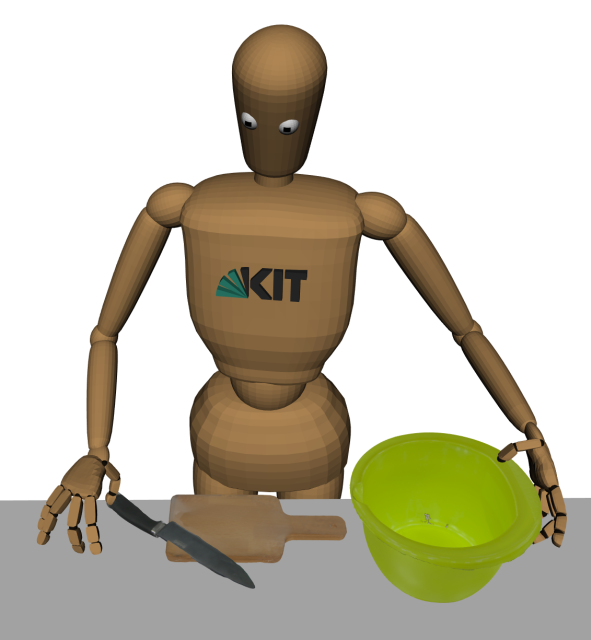}
        \caption{Retreat}%
        \label{fig:experiments:mmm_recording_images:retreat}%
    \end{subfigure}%
    \hfill%
    \begin{subfigure}{0.1712\columnwidth}%
        \centering
        \includegraphics[height=2.75cm]{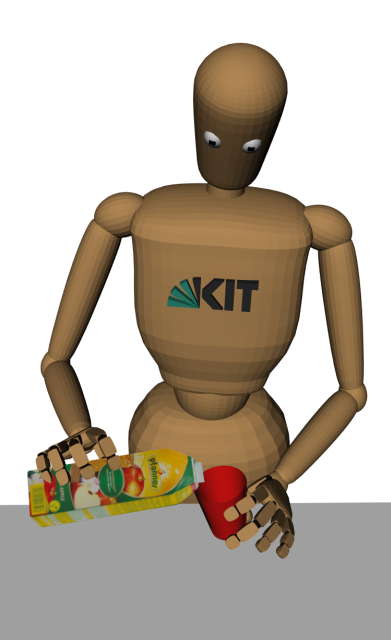}%
        \caption{Pour}%
        \label{fig:experiments:mmm_recording_images:pour}%
    \end{subfigure}%
    \caption{Snapshots of the tasks from motion capture recordings from~\cite{krebsMeixner2021manipulation} that are incrementally learned in Section~\ref{subsec:experiments:learn_from_dataset}.}
    \label{fig:experiments:mmm_recording_images}
    \vspace{-0.5cm}
\end{figure}

We evaluate our approach by incrementally learning a VMP library from human motions recordings.
Specifically, we use 6D hand pose trajectories from selected motions of the KIT Bimanual Actions Dataset~\cite{krebsMeixner2021manipulation}, as shown in Figure~\ref{fig:experiments:mmm_recording_images}. We focus on evaluating the execution of the spatial operations formulated in Section~\ref{sec:FullPoseVMP}. Therefore, we assume to be given which fundamental operation should be performed when. 

\subsection{Incremental Learning of a VMP Library}
\label{subsec:experiments:learn_from_dataset}

First, we evaluate \emph{adding}, \emph{improving}, \emph{merging two modes} and \emph{splitting a mode}. \emph{Removing} is considered as trivial.

\begin{figure}
    \centering
    \input{img/learning_0__incremental_update/approach_combined.pgf}
    \caption{\emph{Added} and incrementally \emph{improved} VMP for an $\mathsf{approach}$ task.
    \emph{Left}: Incrementally provided demonstrations.
    \emph{Middle}: Incrementally learned full-pose VMP, with positions $\bm{p}$ (%
            {\color{matplotlib_red}$\blacksquare$} $x$,
            {\color{matplotlib_green}$\blacksquare$} $y$,
            {\color{matplotlib_blue}$\blacksquare$} $z$%
        ),
        and orientations $\bm{q}$ (%
            {\color{matplotlib_red}$\blacksquare$} $q_x$,
            {\color{matplotlib_green}$\blacksquare$} $q_y$,
            {\color{matplotlib_blue}$\blacksquare$} $q_z$,
            {\color{matplotlib_orange}$\blacksquare$} $q_w$%
        ). All executions are performed w.r.t. the start and end of the first demonstration.
        Colors go from transparent to opaque to show the incremental updates.
    \emph{Right}: Weight means after $6$ demonstrations.
    }
    \label{fig:exp:dataset:add_improve}
    \vspace{-0.05cm}
\end{figure}

\begin{table}
    \centering
    \begin{tabular}{lll}
        \toprule
        Operation & Distance (translation) & Distance (rotation) \\
        \cmidrule(l){1-1} \cmidrule(l){2-3}
        Add and improve & \SI{4.6e-15}{\milli\meter} & below detection limit \\
        Merge two modes & \SI{4.0e-15}{\milli\meter} & \SI{5.4e-7}{\degree} \\
        Split a mode & \SI{22.5}{\milli\meter} and \SI{39.0}{\milli\meter} & \SI{3.2}{\degree} and \SI{6.7}{\degree} \\
        (Without splitting) & \SI{34.1}{\milli\meter} and \SI{68.2}{\milli\meter} & \SI{4.9}{\degree} and \SI{9.9}{\degree} \\
        \bottomrule
    \end{tabular}
    \caption{Average root mean square distances between batch-wise and incremental estimations of the VMP weight means.
    \emph{Without splitting} is displayed for comparison with \emph{split a mode}.
    }
    \label{tab:experiments:distances}
    \vspace{-0.5cm}
\end{table}

\subsubsection{Adding and Improving}
\label{subsec:experiments:addImprove}
We use one demonstration of $\mathsf{approaching}$ an object with the right hand to perform a $\mathsf{sweeping}$ task to create a VMP and \emph{add} it to the library. The VMP is then \emph{improved} incrementally with additional demonstrations.
Figure~\ref{fig:exp:dataset:add_improve} shows the incrementally-provided demonstrations, the incrementally-learned VMPs, and the resulting weights. 
As intended, the addition and improvement operations learn a generalized representation of the demonstrations. 
As expected, quantitative differences between the incremental and batch-wise estimation are negligible (see  Table~\ref{tab:experiments:distances}) since the calculations are mathematically equivalent.

\subsubsection{Merging Modes}
\label{subsec:experiments:mergeModes}
\begin{figure}
    \centering
    \input{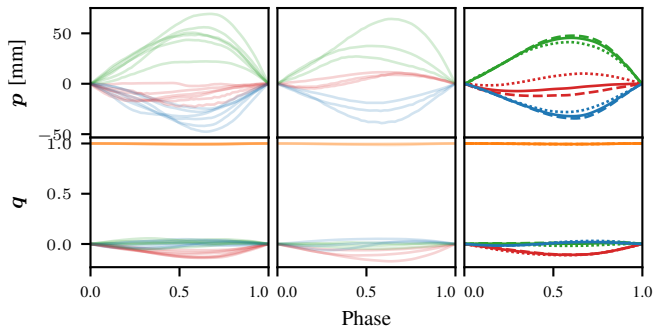}
    \caption{%
        Merge estimations of $\mathsf{lift}$ VMPs. The trajectories are normalized to the same start and end.
        \emph{Left}: Demonstrations in the context of $\mathsf{cutting}$.
        \emph{Middle}: Demonstrations in the context of $\mathsf{peeling}$.
        \emph{Right}: Execution of the individual VMPs from the context of $\mathsf{cutting}$ (- -) and $\mathsf{peeling}$ ($\cdots$), and of the merged VMP (---).
    }
    \label{fig:exp:dataset:merge_modes}
\end{figure}

Next, we consider two VMPs already encoded in the library, namely $\mathsf{lifting \; to \; cut}$ and $\mathsf{lifting \; to \; peel}$, previously trained from 6 and 3 demonstrations, respectively. As shown in Figure~\ref{fig:exp:dataset:merge_modes}-\emph{left}, \emph{middle}, the two VMPs represent similar motions
and are incrementally merged into a single $\mathsf{lifting}$ VMP (see Figure~\ref{fig:exp:dataset:merge_modes}-\emph{right}).
As shown in Table~\ref{tab:experiments:distances}, mathematical equivalence results in negligible differences w.r.t. batch-wise estimations.
The imbalanced number of demonstrations per mode is not detrimental.

\subsubsection{Splitting a Mode}
\label{subsec:experiments:splitMode}

\begin{figure}
    \centering
    \input{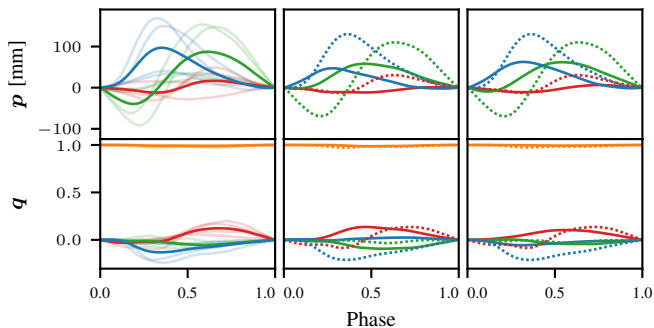}
    \caption{%
        Splitting a $\mathsf{retreat}$ VMP into two modes. The trajectories are normalized to the same start and end.
        \emph{Left}: Joint estimation (---) after 4 demonstrations. Demonstrations and estimated VMP are depicted by semitransparent and opaque lines, respectively.
        \emph{Middle}: Mode 1 (---) and Mode 2 ($\cdots$) after splitting with a $5$th demonstration.
        \emph{Right}: Mode 1 (---) and Mode 2 ($\cdots$) after further improvement from $4$ additional demonstrations.%
    }
    \label{fig:exp:dataset:split_mode}
    \vspace{-0.5cm}
\end{figure}

We now study the $\mathsf{retreat}$ segments of $\mathsf{transfer}$ motions, for which the single VMP encoded in the library is not sufficient. As shown in Figure~\ref{fig:exp:dataset:split_mode}-\emph{left}, the learned $\mathsf{retreat}$ VMP is not able to represent the elements $y$ and $q_z$ of the $4$ demonstrated full-pose trajectories. Therefore, the $5$th demonstration is used to split the learned VMP into two modes. Figure~\ref{fig:exp:dataset:split_mode}-\emph{middle} shows the resulting two modes, which result in better representations of the demonstrations.
$4$ additional demonstrations, each automatically assigned to the more probable mode, are further leveraged to refine the VMPs (see Figure~\ref{fig:exp:dataset:split_mode}-\emph{right}).
As shown in Table~\ref{tab:experiments:distances}, the incremental estimation of splitting a mode differs from batch-wise estimations computed from the demonstrations of the respective modes.
However, the differences are still reasonable, and lower than for a single VMP learned from all demonstrations.
Therefore, the incremental splitting was successful, despite the assumption of both modes having been observed equally often at the time of splitting being violated ($3$ and $2$ observations) and the overall number of demonstrations per mode ($6$ and $3$) being imbalanced.

\subsection{Via-Point Estimation}
\label{subsec:experiments:viaPoints}

\begin{figure}
    \centering
    \input{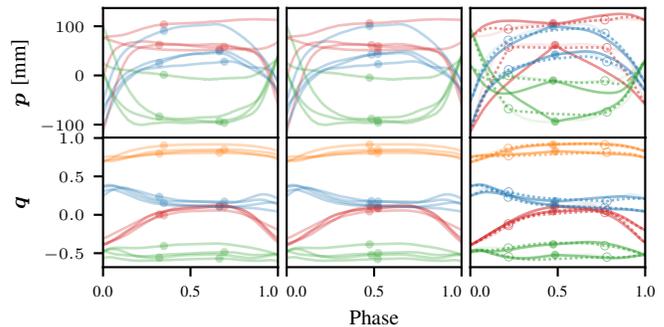}
    \caption{%
        Estimation of via-points for $9$ demonstrations of $\mathsf{pouring}$ actions. For clarity, only $4$ (left, middle) and $2$ of the demonstrations (right) are visualized.
        \emph{Left:} Detection of a via-point using \emph{maximum distance}.
        \emph{Middle:} Detection of a via-point using \emph{weighted phase average}.
        \emph{Right:} Execution using $1$ (---/$\bullet$) or $2$ ($\cdots$/$\circ$) via-points detected by \emph{weighted phase average}.
    }
    \label{fig:exp:dataset:via_points}
\end{figure}

Next, we compare the three approaches for via-point detection presented in Section~\ref{subsec:detectionOfViaPoints} on demonstrations of a $\mathsf{pouring}$ task, whose amplitudes differ strongly. At first, we detect one via-point per demonstration.
As shown in Figure~\ref{fig:exp:dataset:via_points}-\emph{left}, the \emph{maximum weight} approach leads to strongly-varying via-points across the demonstrations. Instead, the \emph{weighted phase average} (\emph{middle}) yields more consistent results, that are better suited to generalize the estimation.
As shown in
Table~\ref{tab:experiments:viaPointDistancesAndDurations}, the
\emph{maximum weight} and \emph{brute force} approaches result in similar distances between demonstration and reconstruction with a high estimation time for the latter. 
The \emph{weighted phase average} is competitively fast, while resulting in similar distances.
Figure~\ref{fig:exp:dataset:via_points}-\emph{right} shows examples of reconstructions obtained with up to $1$ or $3$ via-points detected via \emph{weighted phase average}.
Here, the $3$ points are reduced to $2$ in the removal step.
For this motion, a single via-point is not sufficient as only the midpart of the reproduction matches the demonstration. 
Trajectories obtained with pairs of via-points reproduces the demonstrations considerably better.

\begin{table}
    \centering
    \begin{tabular}{llll}
        \toprule
        \# & Approach & Average Distance & Duration [s] \\
        \cmidrule(l){1-2} \cmidrule(l){3-3} \cmidrule(l){4-4}
            0 & None & \SI{53.8+-22.8}{} & \hspace{6.5ex}--- \\
            \multirow{3}{*}{1\vspace{2em}} & \emph{Maximum Distance} & \SI{20.5+-10.8}{} & \hspace{1ex}\SI{0.34+-0.03}{} \\
            & \emph{Brute Force} & \SI{20.9+-10.4}{} & \SI{81.85+-12.50}{} \\
            & \emph{Weighted Phase Average} & \SI{22.6+-9.3}{} & \hspace{1ex}\SI{0.33+-0.03}{} \\
            \multirow{2}{*}{3\vspace{1em}} & \emph{Maximum Distance} & \SI{10.2+-2.9}{} & \hspace{1ex}\SI{3.11+-0.40}{} \\
            & \emph{Weighted Phase Average} & \SI{10.1+-3.4}{} & \hspace{1ex}\SI{2.79+-0.87}{} \\
        \bottomrule
    \end{tabular}
    \caption{Evaluation of via-point detection approaches. Average weighted distance between demonstrations and VMP reconstructions and detection duration. \emph{Brute force} is omitted for $3$ via-points for runtime reasons.
    \#~denotes the maximum number of via-points to be detected.}
    \label{tab:experiments:viaPointDistancesAndDurations}
    \vspace{-0.2cm}
\end{table}

\section{Conclusion and Future Work}

In this paper, we identified seven fundamental operations to incrementally learn MP libraries.
We formulated all spatial operations for a specific type of MP library, thus providing a framework to incrementally learn VMPs, detect their via-points, and estimate their task parameters.
We achieve mathematical equivalence to batch-wise estimation of VMP weights for all but one of them and convergence for via-points and task parameters.
Our approach soundly treats full-pose trajectories by building on Riemannian manifolds theory. Importantly, it follows a strict definition of incremental learning that limits the amount of stored information.

\addtolength{\textheight}{-7.5cm}
In general, incremental learning of MP libraries opened the three main challenges of \emph{how} and \emph{when} to apply \emph{which} of the presented fundamental operations. We see the first of these challenges, tackled in this paper, as a requirement to address the latter ones. Future work will consist in deciding \emph{when} to incrementally apply \emph{which} operation to learn generalizable MP libraries from a large set of demonstrations. 
Once solutions to all three challenges are sufficiently mature, the full potential of incremental learning might be leveraged for long-term usage of MP libraries, thus raising novel challenges regarding the stability and convergence of such a library.
Moreover, we envision that the presented solutions on \emph{how} to apply the fundamental operations can be transferred to other MPs building on basis functions, such as DMPs and ProMPs. 
In addition, we will treat the temporal operations, which requires reliable segmentation approaches.

\clearpage
\bibliographystyle{ieeetr}
\bibliography{references.bib}

\begin{thebibliography}{10}

\bibitem{billard2008handbookOfRobotics}
A.~Billard, S.~Calinon, R.~Dillmann, and S.~Schaal, ``Robot {{Programming}} by {{Demonstration}},'' in {\em Handbook of {{Robotics}}}, pp.~1371--1394, {Springer}, 2008.

\bibitem{schaal1999route}
S.~Schaal, ``Is {{Imitation Learning}} the {{Route}} to {{Humanoid Robots}}?,'' {\em Trends in Cognitive Sciences}, vol.~3, no.~6, pp.~233--242, 1999.

\bibitem{Giszter1993}
S.~F. Giszter, F.~A. Mussa-Ivaldi, and E.~Bizzi, ``Convergent {{Force Fields Organized}} in the {{Frog’s Spinal Cord}},'' {\em Journal of Neuroscience}, vol.~13, pp.~467--491, 1993.

\bibitem{pastor2009motorSkills}
P.~Pastor, H.~Hoffmann, T.~Asfour, and S.~Schaal, ``Learning and {{Generalization}} of {{Motor Skills}} by {{Learning From Demonstration}},'' in {\em {IEEE} Intl. Conf. on Robotics and Automation ({ICRA})}, pp.~763--768, 2009.

\bibitem{GepperthHammer2016incremental}
A.~Gepperth and B.~Hammer, ``Incremental {{Learning}} {{Algorithms}} {{and Applications}},'' in {\em European {{Symposium}} on {{Artificial Neural Networks}} ({{ESANN}})}, 2016.

\bibitem{losingIncrementalOnlineLearning2018}
V.~Losing, B.~Hammer, and H.~Wersing, ``Incremental {{On-Line Learning}}: {{A}} {{Review}} and {{Comparison}} of {{State}} of the {{Art Algorithms}},'' {\em Neurocomputing}, vol.~275, pp.~1261--1274, 2018.

\bibitem{zhou2019vmp}
Y.~Zhou, J.~Gao, and T.~Asfour, ``Learning {{Via-Point Movement Primitives With Inter-}} and {{Extrapolation Capabilities}},'' in {\em {IEEE/RSJ} Intl. Conf. on Intelligent Robots and Systems ({IROS})}, pp.~4301--4308, 2019.

\bibitem{kulic2012primitiveTree}
D.~Kuli{\'c}, C.~Ott, D.~Lee, J.~Ishikawa, and Y.~Nakamura, ``Incremental {{Learning}} of {{Full Body Motion Primitives}} and {{Their Sequencing Through Human Motion Observation}},'' {\em Intl. Journal of Robotics Research}, vol.~31, no.~3, pp.~330--345, 2012.

\bibitem{niekum2013fsa}
S.~Niekum, S.~Chitta, A.~Barto, B.~Marthi, and S.~Osentoski, ``Incremental {{Semantically Grounded Learning From Demonstration}},'' in {\em Robotics: Science and Systems ({R:SS})}, 2013.

\bibitem{niekumLearningGroundedFiniteState2015}
S.~Niekum, S.~Osentoski, G.~Konidaris, S.~Chitta, B.~Marthi, and A.~G. Barto, ``Learning {{Grounded Finite-State Representations From Unstructured Demonstrations}},'' {\em Intl. Journal of Robotics Research}, vol.~34, no.~2, pp.~131--157, 2015.

\bibitem{ijspeert2013dmp}
A.~J. Ijspeert, J.~Nakanishi, H.~Hoffmann, P.~Pastor, and S.~Schaal, ``Dynamical {{Movement}} {{Primitives}}: {{Learning}} {{Attractor}} {{Models}} for {{Motor}} {{Behaviors}},'' {\em Neural Computation}, vol.~25, no.~2, pp.~328--373, 2013.

\bibitem{gutierrezIncrementalTaskModification2018}
R.~A. Gutierrez, V.~Chu, A.~L. Thomaz, and S.~Niekum, ``Incremental {{Task Modification}} via {{Corrective Demonstrations}},'' in {\em {IEEE} Intl. Conf. on Robotics and Automation ({ICRA})}, pp.~1126--1133, 2018.

\bibitem{gutierrezLearningCorrectiveDemonstrations2019}
R.~A. Gutierrez, E.~S. Short, S.~Niekum, and A.~L. Thomaz, ``Learning {{From Corrective Demonstrations}},'' in {\em {ACM/IEEE} Intl. Conf. on Human-Robot Interaction ({HRI})}, pp.~712--714, 2019.

\bibitem{meier2012regonition}
F.~Meier, E.~Theodorou, and S.~Schaal, ``Movement {{Segmentation}} and {{Recognition}} for {{Imitation Learning}},'' in {\em International {{Conference}} on {{Artificial Intelligence}} and {{Statistics}} ({{AISTATS}})}, pp.~761--769, 2012.

\bibitem{gamsAdaptationCoachingPeriodic2016}
A.~Gams, T.~Petri{\v c}, M.~Do, B.~Nemec, J.~Morimoto, T.~Asfour, and A.~Ude, ``Adaptation and {{Coaching}} of {{Periodic}} {{Motion}} {{Primitives}} through {{Physical}} and {{Visual}} {{Interaction}},'' {\em Robotics and Autonomous Systems}, vol.~75, pp.~340--351, 2016.

\bibitem{calinonIncrementalLearningGestures2007}
S.~Calinon and A.~Billard, ``Incremental {{Learning}} of {{Gestures}} by {{Imitation}} in a {{Humanoid Robot}},'' in {\em {ACM/IEEE} Intl. Conf. on Human-Robot Interaction ({HRI})}, p.~255, 2007.

\bibitem{kronanderIncrementalMotionLearning2015}
K.~Kronander, M.~Khansari, and A.~Billard, ``Incremental {{Motion}} {{Learning}} with {{Locally}} {{Modulated}} {{Dynamical}} {{Systems}},'' {\em Robotics and Autonomous Systems}, vol.~70, pp.~52--62, 2015.

\bibitem{takano2016incrementalHMM}
W.~Takano and Y.~Nakamura, ``Real-{{Time Unsupervised Segmentation}} of {{Human Whole-Body Motion}} and {{Its Application}} to {{Humanoid Robot Acquisition}} of {{Motion Symbols}},'' {\em Robotics and Autonomous Systems}, vol.~75, pp.~260--272, 2016.

\bibitem{pastorAssociativeSkillMemories2012}
P.~Pastor, M.~Kalakrishnan, L.~Righetti, and S.~Schaal, ``Towards {{Associative Skill Memories}},'' in {\em {IEEE/RAS} Intl. Conf. on Humanoid Robots (Humanoids)}, pp.~309--315, 2012.

\bibitem{lemme2014bootstrapping}
A.~Lemme, R.~F. Reinhart, and J.~J. Steil, ``Self-{{Supervised Bootstrapping}} of a {{Movement Primitive Library From Complex Trajectories}},'' in {\em {IEEE/RAS} Intl. Conf. on Humanoid Robots (Humanoids)}, pp.~726--732, 2014.

\bibitem{paraschos2013promp}
A.~Paraschos, C.~Daniel, J.~R. Peters, and G.~Neumann, ``Probabilistic {{Movement Primitives}},'' in {\em Neural Information Processing Systems ({NeurIPS})}, vol.~26, pp.~2616--2624, 2013.

\bibitem{rozoOrientationProbabilisticMovement2021}
L.~Rozo and V.~Dave, ``Orientation {{Probabilistic Movement Primitives}} on {{Riemannian Manifolds}},'' in {\em Conference on Robot Learning ({CoRL})}, 2021.

\bibitem{Zhang23:RiemannianStableDS}
J.~Zhang, H.~B. Mohammadi, and L.~Rozo, ``Learning {R}iemannian stable dynamical systems via diffeomorphisms,'' in {\em Conference on Robot Learning ({CoRL})}, pp.~1211--1221, 2022.

\bibitem{Lee13:SmoothManifolds}
J.~M. Lee, {\em {Introduction to smooth manifolds}}.
\newblock Springer, 2013.

\bibitem{Lee18:RiemannianManifolds}
J.~M. Lee, {\em Introduction to {R}iemannian Manifolds}.
\newblock Springer, 2018.

\bibitem{fletcher2013geodesicRegression}
P.~Thomas~Fletcher, ``Geodesic {{Regression}} and the {{Theory}} of {{Least Squares}} on {{Riemannian Manifolds}},'' {\em Intl. Journal on Computer Vision}, vol.~105, no.~2, pp.~171--185, 2013.

\bibitem{kim2014multivariateLinearRegression}
H.~J. Kim, N.~Adluru, M.~D. Collins, M.~K. Chung, B.~B. Bendin, S.~C. Johnson, R.~J. Davidson, and V.~Singh, ``Multivariate {{General Linear Models}} ({{MGLM}}) on {{Riemannian Manifolds}} with {{Applications}} to {{Statistical Analysis}} of {{Diffusion Weighted Images}},'' in {\em Conf. on Computer Vision and Pattern Recognition ({CVPR})}, pp.~2705--2712, 2014.

\bibitem{pennec2006basicTools}
X.~Pennec, ``Intrinsic {{Statistics}} on {{Riemannian Manifolds}}: {{Basic Tools}} for {{Geometric Measurements}},'' {\em Journal of Mathematical Imaging and Vision}, vol.~25, no.~1, pp.~127--154, 2006.

\bibitem{salehian2015ifme}
H.~Salehian, R.~Chakraborty, E.~Ofori, D.~Vaillancourt, and B.~C. Vemuri, ``An {{Efficient}} {{Recursive}} {{Estimator}} of the {{Fr\'{e}chet}} {{Mean}} on a {{Hypersphere}} with {{Applications}} to {{Medical Image Analysis}},'' {\em Mathematical Foundations of Computational Anatomy}, vol.~3, pp.~143--154, 2015.

\bibitem{krebsMeixner2021manipulation}
F.~Krebs, A.~Meixner, I.~Patzer, and T.~Asfour, ``The {{KIT Bimanual Manipulation Dataset}},'' in {\em {IEEE/RAS} Intl. Conf. on Humanoid Robots (Humanoids)}, pp.~499--506, 2021.

\end{thebibliography}

\end{document}